\pdfoutput=1

\documentclass[11pt]{article}

\usepackage{naacl2021}
\usepackage{graphicx}
\usepackage{times}
\usepackage{latexsym}
\usepackage{amsmath}
\usepackage{booktabs}
\usepackage{pifont}
\newcommand{\cmark}{\ding{51}}%
\newcommand{\xmark}{\ding{55}}%
\usepackage[T1]{fontenc}

\usepackage[utf8]{inputenc}

\usepackage{microtype}

\title{\textsc{Deux}: An Attribute-Guided Framework for\\
Sociable Recommendation Dialog Systems}

\author{$^{\dagger}$Yu Li \quad $^{\ddagger}$Shirley Anugrah Hayati \quad  $^{\mathsection}$Weiyan Shi \quad $^{\mathsection}$Zhou Yu\\
~~\\
$^{\dagger}$Department of Computer Science, University of California, Davis\\
$^{\ddagger}$University of Pennsylvania\\
$^{\mathsection}$Department of Computer Science, Columbia University\\
  $^{\dagger}$\texttt{yooli@ucdavis.edu}, $^{\ddagger}$\texttt{sahayati@upenn.edu}\\
  $^{\mathsection}$\texttt{\{ws2634, zy2461\}@columbia.edu} \\}

\begin{document}
\maketitle
\begin{abstract}
It is important for sociable recommendation dialog systems to perform as both on-task content and social content to engage users and gain their favor. In addition to understand the user preferences and provide a satisfying recommendation, such systems must be able to generate coherent and natural social conversations to the user. Traditional dialog state tracking cannot be applied to such systems because it does not track the attributes in the social content.
To address this challenge, we propose \textsc{Deux}, a novel attribute-guided framework to create better user experiences while accomplishing a movie recommendation task. \textsc{Deux} has a module that keeps track of the movie attributes (e.g., favorite genres, actors, etc.) in both user utterances and system responses. This allows the system to introduce new movie attributes in its social content. Then, \textsc{Deux} has multiple values for the same attribute type which suits the recommendation task since a user may like multiple genres, for instance. Experiments suggest that \textsc{Deux} outperforms all the baselines on being more consistent, fitting the user preferences better, and providing a more engaging chat experience. Our approach can be used for any similar problems of sociable task-oriented dialog system.


\end{abstract}

\section{Introduction}
\begin{figure}[t!]
\vspace{0mm}
\centering
{
\includegraphics[trim={0 0 0 0},clip,width=7.7cm]{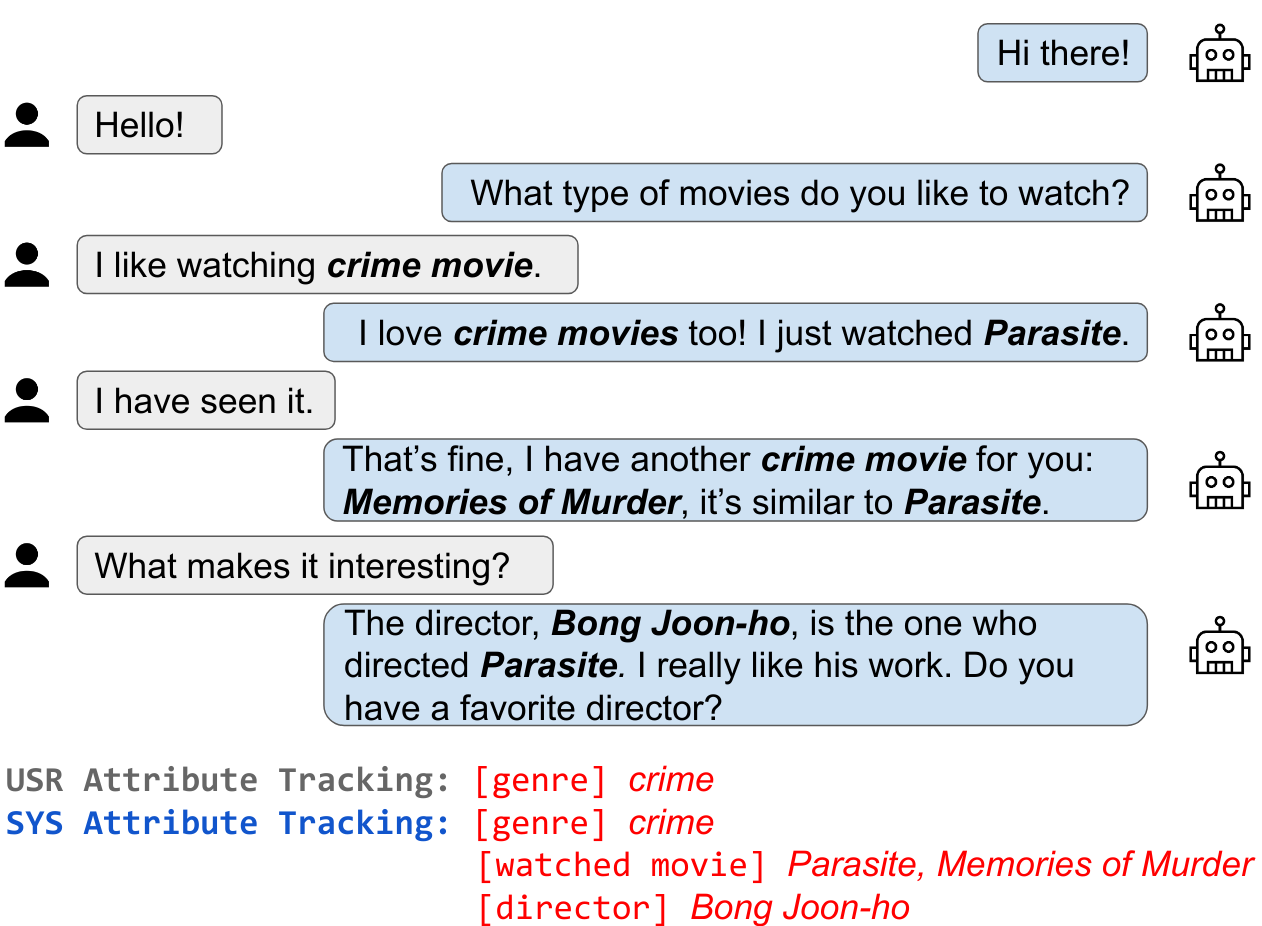}
}
\caption{\label{fig:example_chat}Our framework, \textsc{Deux}, can handle the user preferences of crime and recommend another similar crime movie because it tracks the attributes in both previous user utterances and system responses.}
\vspace{0mm}
\end{figure}


In recent years, research in recommendation dialog systems have received increasing interest. They commonly focus on two approaches: task-oriented dialog systems or free-form dialog systems with more diverse interactions. 
Traditional task-oriented recommendation dialog systems collect information by asking questions with predefined slots and limited values \cite{zhang2018towards, sun2018conversational, christakopoulou2018q, lee2018making}. While the response is controllable, such systems can only provide limited and rigid responses, which could neglect the user experience in the conversation. On the other hand, with the increasing demand for engaging personal conversational assistants, an ideal recommendation dialog system should be natural and sociable to gain trust and favor from the users \cite{hayati2020inspired}. 

To address the rigid response problems in task-oriented recommendation systems, previous studies have developed free-form recommendation dialog systems that enable more diverse social interactions and better user experiences \cite{li2018towards, kang2019-recommendation, chen-etal-2019-towards, liu-etal-2020-towards}. These systems do not use predefined slots and make recommendations without slot filling. As a result, they suffer from understanding user preferences correctly and updating the dialog states accurately. Therefore, it is important to combine the strength of the two approaches to build controllable sociable recommendation dialogue systems.

However, it is challenging to track traditional dialog states in free-form recommendation dialogs because (1) \textbf{the on-task content and the social content are intertwined in the conversation} and (2) \textbf{new attributes could be introduced in the social content}. For example, in a movie recommendation setting, the system needs to accomplish the recommendation task with on-task responses, such as asking for a user's favorite genre. Meanwhile, to make the conversation more sociable, the system also needs to respond to the user with social content, such as commenting on a movie. These social content introduces new movie titles and their related attributes (e.g., genres and actors) and drives the conversation forward. Thus, it is difficult to define specific dialog states and update dialog policies in these free-form dialog systems. As a result, although these systems can generate fluent responses, they are often redundant and inappropriate. 

To address these challenges, we propose \textsc{Deux}, a novel attribute-guided framework to create better user experiences while accomplishing the recommendation task. \textsc{Deux} has a module that keeps track of the attributes (e.g., favorite genres, actors, etc.) in both user utterances and system responses. Instead of providing a response solely based on the attributes already mentioned in the context, we also have a system attribute predictor module to predict the preferred attributes in the next system response. Finally, the response is conditionally generated on the predicted preferred attributes, leading to a more controllable generation compared to the existing free-form recommendation dialog systems. As shown in Figure \ref{fig:example_chat}, our response is more engaging and consistent than the responses of both a task-oriented system and an existing free-form recommendation dialog system. 

\begin{table}
    \centering
    \begin{tabular}{lcc}
    \toprule
    \textbf{Characteristics} & \textbf{\textsc{Deux}} &  \textbf{DST} \\
    \toprule
   Tracking user's preference &  \cmark &  \cmark\\
   Tracking system's preference & \cmark &  \xmark \\
   Multiple slot values & \cmark &  \xmark\\
    \bottomrule
    \end{tabular}
    \caption{Comparison between our framework \textsc{Deux} and traditional dialog state tracking (DST).}
    \label{tab:comparison}
    \vspace{-5mm}
\end{table}

Our main contribution is as follows. We introduce a new definition of attribute tracking in the sociable recommendation task. Our attribute tracking has two main differences with the dialog states in traditional task-oriented dialog systems as shown in Table \ref{tab:comparison}. Unlike traditional dialog states which only focus on the user's preference, \textsc{Deux} includes attributes in previous system responses. These additional attributes are about system preferences because the system should also introduce its own opinion and new attributes in sociable recommendation tasks. On the other hand, while traditional dialog states also track attributes on the system's side, that tracking is only used when the dialog system requests the user's preference. Then, instead of limiting only one value for one slot type, we have multiple values for the same slot type. For instance, the slot type of movie genre may contain actions, comedies, and dramas at the same time.

Note that a complete recommendation dialog system includes a dialog system that interacts with the user to collect user preferences and a recommender system that converts user preferences into queries and retrieves movies from a database.  In this work, we focus on the dialog system, so any off-shelf recommender systems can be easily plugged into our framework. The advantage of such modular approach is that the dialog system and recommender system can be improved separately. 

We evaluate our framework on \textsc{Inspired}, a sociable movie recommendation dialog dataset that contains human-human conversations with rich social content \cite{hayati2020inspired}. Human evaluations suggest that \textsc{Deux} outperforms multiple competitive baselines on different metrics.

\section{Related Work}

Early research in recommendation dialog systems has utilized task-oriented slot-filling approach. In their conversations, the systems ask questions about user preference to fill in the predefined slots and later select items for recommendation \cite{zhang2018towards, sun2018conversational, christakopoulou2018q, lee2018making}. This method has explicit dialog states because the number of slot types and values is limited. Therefore, they mostly use fixed template-based response generation where the variables are filled in by the generator, resulting in rigid sentences and monotonous conversations. 


\begin{figure*}
    \centering
    \includegraphics[trim={0 0 0 0},clip,width=16cm]{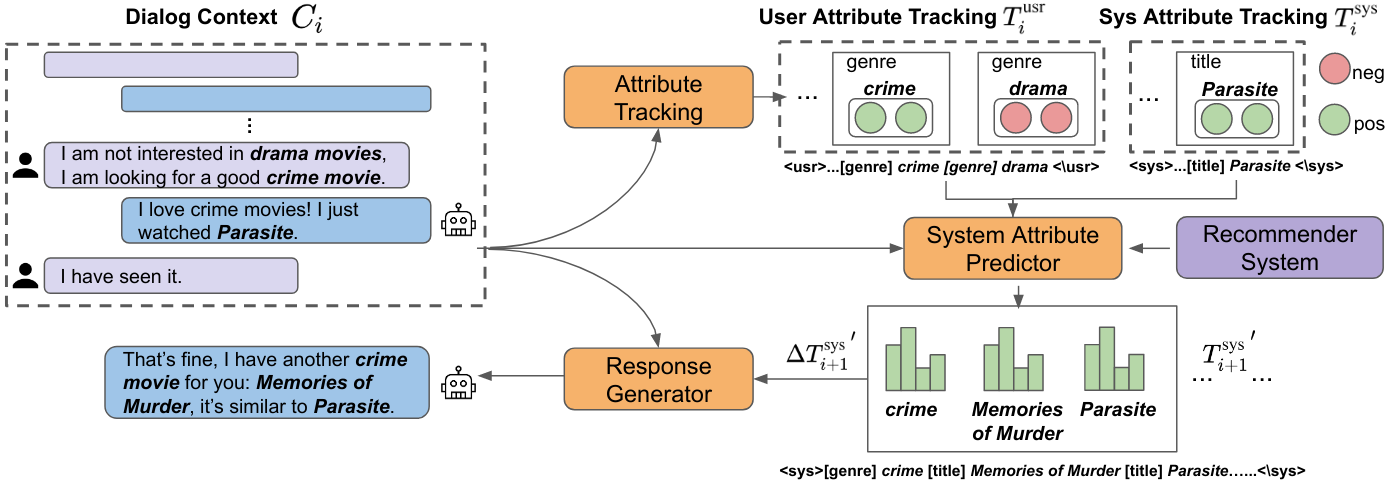}
    \caption{The left panel shows an on-going movie recommendation dialog, for each turn, the attribute tracking module (section \ref{entity_tracking}) takes dialog history as input and predicts user's attribute tracking and system's attribute tracking at current turn. Then the system attribute predictor module (section \ref{system_entity_predictor}) takes both current entity tracking and dialog history as input, outputs at each decoding step to predict the system's attribute tracking at next turn. Finally, the response generator module (section \ref{response_generator}) generates the next response conditionally on the predicted attributes.}
    \label{fig:method}
    \vspace{-3mm}
\end{figure*}

Recently, to make the conversation more natural and engaging, researchers  have shifted toward free-interaction recommendation dialog systems \cite{li2018towards, kang2019-recommendation, moon2019opendialkg, chen-etal-2019-towards, zhou2020improving, hayati2020inspired}. \citet{li2018towards, kang2019-recommendation} develop free-interaction recommendation dialog systems on human-to-human multi-turn movie recommendation dialog datasets. \citet{moon2019opendialkg} create a recommendation dialog dataset with parallel attributes in the knowledge graph. \citet{chen-etal-2019-towards, zhou2020improving} integrate the recommender system and the dialog system to incorporate the knowledge from recommender system into response generation. 

These approaches mostly build the recommender system and the dialog system together. Meanwhile, our approach focuses on the dialog system and any off-shelf recommender system can be plugged into our framework following \citet{hayati2020inspired}. Rather than using a sentiment classifier to detect user preference for a certain attribute as in \citet{hayati2020inspired} and \citet{li2018towards}, for each turn in the conversation, we represent the user preferences as a set of attributes of user preferences from both user utterances and system responses. Furthermore, we predict the attributes that could appeal to the user in the next system response so that we can track the status in the dialog system.

In traditional task-oriented dialog systems, dialog state tracking predicts belief state which contains limited domains, slot types, and slot values in the dialog history \cite{henderson2013deep, mrkvsic2016neural, rastogi2017scalable}. These works only focus on the dialog state tracking module. Recent improvements built on top of the transformer and pre-trained language models \cite{devlin2019bert, liu2019roberta, radford2019language, yang2020xlnet}. Lately,  \citet{hosseini2020simple} have obtained state-of-the-art results of dialogue state tracking on Multi-Domain Wizard-of-Oz dataset \cite{budzianowski2020multiwoz}. Our framework is also developed on top of large scale pre-trained language models. However, we track not only the information in the user's dialog history, but also the attributes that the system delivers in the previous system responses. Since attributes could exist in both on-task and social content, our framework leads to a better combination of the social content and on-task content during the response generation.

To interleave the on-task content with the social content in dialog systems, \citet{yu2017learning, papaioannou2017hybrid} build hybrid dialog systems that combine a task-oriented model and a social model. In these studies, a selector is designed to choose an appropriate output from one of the systems \cite{yu2017learning} and a connector to combine two responses \cite{zhao2018sogo}. \citet{li2020end} proposes a hierarchical intent annotation scheme that divides on-task and off-task content in the dataset and trains a model handling both types of content in non-collaborative tasks. Compared with these efforts, we track the attributes in the history instead of classifying the user and the system's intents. \textsc{Deux}'s approach is appropriate for recommendation task since the system can provide both on-task and social content containing attributes, and finally make a recommendation based on these attributes. Combining both attributes and intents will be an interesting future work.

\section{Our Approach: \textsc{Deux}}
We demonstrate the overall architecture of \textsc{Deux}, our framework for recommendation dialog system, in Figure \ref{fig:method}. 
To enable easy replacement with any off-shelf recommender system, we decouple the system into separate modules: attribute tracking (section \ref{entity_tracking}), system attribute predictor (section \ref{system_entity_predictor}) and response generator (section \ref{response_generator}). 

For each turn, the attribute tracking module predicts user's attribute tracking and system's attribute tracking at current turn. Then the system attribute predictor module predicts the system's attribute tracking at next turn. Finally, the response generator module generates the next response conditionally on predicted attributes.
Suppose we have a recommendation dialog corpus with $N$ turns, as shown in the dialogue context box in Figure \ref{fig:method}:
\begin{align}
    \mathcal{D}=\{(U_i,E_i^{\textrm{usr}},E_i^{\textrm{sys}},T_i^{\textrm{usr}}, T_i^{\textrm{sys}}, Y_i)\}_{i=1}^N
\end{align}
\noindent where $\forall(U_i,E_i^{\textrm{usr}},E_i^{\textrm{sys}},T_i^{\textrm{usr}}, T_i^{\textrm{sys}}, Y_i)\in\mathcal{D}$, and $C_i=(U_{1},\cdots, U_{N})$ represents the dialog history consisting utterance at $i$-th turn. $E_i^{\textrm{usr}}$ is a set of attributes which user mentions in the dialog history. $E_i^{\textrm{sys}}$ is a set of attributes that system mentions in the dialog history. $T_i^{\textrm{usr}}$ and $T_i^{\textrm{sys}}$ are the attribute tracking in user utterances and system responses. $Y_i$ represents the system responses.


\subsection{Attribute Tracking}\label{entity_tracking}
Given the dialog history at each dialog turn $i$, the goal of this module is to predict the attribute tracking $T_i^{\textrm{usr}}$ and $T_{i}^{\textrm{sys}}$. Specifically, we aim to extract the attributes that a user prefers in both user utterances and system responses. Then the system will recommend the item that can meet the preference. We use a positive or negative label to categorize attributes in the attribute tracking\footnote{Section \ref{data_preprocessing} explains more details about how we obtain these positive and negative attributes during training.}. A positive attribute means the user prefer a recommendation that is related to this attribute while a negative attribute means the user has relatively low interest to this attribute. $T_i^{\textrm{usr}}$ and $T_{i}^{\textrm{sys}}$ can be denoted as:
\begin{align}
    &\begin{aligned}
        T_i^{\textrm{usr}}=\{(e_{i,j}^{\textrm{usr}}, l_{i,j}^{\textrm{usr}})\,|\, e_{i,j}^{\textrm{usr}}&\in E_i^{\textrm{usr}},\\ l_{i,j}^{\textrm{usr}}&\in\{\textrm{pos},\textrm{neg}\}\}
    \end{aligned}\\
    &\begin{aligned}
        T_i^{\textrm{sys}}=\{(e_{i,j}^{\textrm{sys}}, l_{i,j}^{\textrm{sys}})\,|\, e_{i,j}^{\textrm{sys}}&\in E_i^{\textrm{sys}},\\ l_{i,j}^{\textrm{sys}}&\in\{\textrm{pos},\textrm{neg}\}\}
    \end{aligned}
\end{align}
where j is the number of attributes in $E_i$.


In our architecture, to detect positive attributes in the dialog history, we first use a public named-entity recognition tool from \citet{gunrock2020} and regular expressions to extract all the movie-related attributes (e.g., movie genres, movie titles and person's names). Then we build a classifier on top of BERT \cite{devlin2019bert} to predict the label for each attribute due to BERT's predominant performance on text classification tasks. This process can be formally denoted as: 
\begin{align}
    [T_i^{\textrm{usr}}, T_i^{\textrm{sys}}]= \textrm{ET}([C_i, E_i^{\textrm{usr}},E_i^{\textrm{sys}}])
\end{align}
\noindent where $\textrm{ET}$ represents the entity tracking module.

\subsection{System Attribute Predictor}\label{system_entity_predictor}
This module aims to predict the appropriate attributes that the system could mention in the response. Therefore, the system can actively drive the conversation forward as well as provide satisfying recommendation. Specifically, its input consists of user's attribute tracking $T_i^{\textrm{usr}}$, system's attribute tracking $T_i^{\textrm{sys}}$ and the dialog history $C_i$. Then, it predicts the system's attribute tracking ${T_{i+1}^{\textrm{sys}}}'$ at next turn. 

As mentioned above, the system attribute tracking module consists of attributes in previous system responses and their sentiment labels. Meanwhile, the relations between these attributes, such as the relation between the actors and the movies, are mainly from the recommender system. 
Since we only focus on the dialog system, our module for predicting the system attribute consists of these three steps: 
\begin{enumerate}
    \item We replace the attributes in current turn attribute tracking with placeholders as the multi-slots. 
    \item Then, we build a system attribute predictor model that takes a set of placeholders as input and generates a a set of positive placeholders in the predicted system attribute tracking.
    \item Finally, we use the plugged recommender system to replace the placeholders with relative attributes and output the final predicted system attribute tracking.
\end{enumerate}

Our system attribute predictor model is adapted from the generative pre-trained language model (GPT-2) \cite{radford2019language}. We fine-tune our model with delexicalized text. During training, we concatenate dialog history and indexed positive placeholders as the input. Meanwhile, in the output, because there could be placeholders that are not in the input dialog history, we substitute all the placeholders that have not been observed in the dialog history with a special ``\texttt{new\_attribute}'' token. Later, we replace all the ``\texttt{new\_attribute}'' tokens and placeholders with the attributes of the recommended items provided by the plugged recommender system.
We formulate the process in this module as:
\begin{align}
    {T_{i+1}^{\textrm{sys}}}' = \textrm{Rel}(\textrm{SEP}(\textrm{Del}([C_i, T_i^{\textrm{usr}}, T_i^{\textrm{sys}}])))
\end{align}
\noindent where $\textrm{Del}$ is the delexicalization process and $\textrm{Rel}$ is the relexicalization process, $\textrm{SEP}$ represents the system attribute predictor model.

\subsection{Response Generator}\label{response_generator}

Our final module aims to produce a response that is conditioned on the predicted system attribute tracking for the next turn by using the output from the system attribute predictor. More concretely, we calculate the attribute difference between the current system attribute tracking and the next system attribute tracking $\Delta {T_i^{s}}'$ as:

\begin{equation}
    \Delta {T_i^{\textrm{sys}}}'=\{e_{i,j}|e_{i,j}\in {T_{i+1}^{\textrm{sys}}}', e_{i,j}\notin T_i^{\textrm{sys}}\}
\end{equation}
Attributes in $\Delta {T_i^{\textrm{sys}}}'$ are considered to be the appropriate attributes that should be in the system response. Thus, we concatenate the dialog history $C_i$ and $\Delta {T_i^{\textrm{sys}}}'$ as the input of the response generator:
\begin{align}
    Y_i =\textrm{RG}([C_i, \Delta {T_i^{\textrm{sys}}}'])
\end{align}
\noindent where $\textrm{RG}$ is the response generator module. We fine-tune the BlenderBot model on \textsc{Inspired} dataset as our response generator. We choose BlenderBot as our base because it is the state-of-the-art model for open-domain chatbot \cite{roller2020recipes}.

\subsection{Data Preprocessing}
\label{data_preprocessing}
To automatically annotate the positive and negative labels for each attribute in attribute tracking, we extract all the attributes and movie recommendations in the dialogs. Then we search for every recommended movie, its genres, actors, and directors in the movie database.

If the recommended movie has a relation to an attribute which is mentioned in the previous dialog history, we annotate this attribute as a positive attribute. Otherwise, we annotate the attribute as a negative attribute. Note that there are commonly multiple recommended items through the conversation in a free-form recommendation dialog dataset. This is because in a sociable recommendation dialog, the user discloses more favor and disfavors. As the system elicits more information of user preferences, users may reject the movie recommendation and ask for a different movie. This changes a positive attribute to negative for the next recommendation. As a result, the same attribute could have different labels in different dialog turns. Thus, we need annotate the attributes in the dialog history every turn rather than the system's internal state, which means the annotation should be considered with the dialog context together.

\section{Experiments}

\subsection{Dataset}
We evaluate on \textsc{Inspired}, a newly released dataset of movie recommendation dialogs \cite{hayati2020inspired}.  It contains 1,011 human-human conversations with 10.73 average turns with concrete success measures. The dialogs are collected in natural setting where human recommenders are informed with sociable strategies and make recommendations to the seeker who is looking for a movie recommendation. Thus, the dialogs contain diverse social interactions from both recommenders and users, including movie discussions, chit-chat, and encouragement. The dataset is also already automatically annotated with movie attribute placeholders (movie titles, actors, and genres), and we use this information for our model.

\textsc{Inspired} contains 1,011 conversations, but there are over 17,000 movies in our movie database. Since our response generator is conditional on the predicted attributes in the system response, we exploit data augmentation to alleviate the sparse data issue when we train the response generator.
Specifically, we first extract all the attributes in the dialog, we then search for the relations between all the attributes in a same dialog. To augment the dataset with more movies, we replace the attributes in a same dialog with another set of attributes which have same relations in the database. Finally, we expand \textsc{Inspired} dataset to 10,000 conversations with different attributes so that the augmented dataset can include more movies.

\subsection{Baselines}
For our baselines, we choose the following dialog models:
\paragraph{\textsc{Inspired} Bot} This is the strategy-based model from \citet{hayati2020inspired}. It utilizes two separate Transformer-based GPT-2 language models  \cite{vaswani2017attention, radford2019language, wu2019alternating} and heuristics to obtain recommendations from off-the-shelf recommender system\footnote{https://www.themoviedb.org/}. 
    
\paragraph{BlenderBot} We fine-tune the BlenderBot model \cite{roller2020recipes} with augmented \textsc{Inspired} dataset as our second baseline. This serves as a baseline for social chatbot since we are not adding recommender system to this baseline. 
    
\paragraph{Hybrid} Following \citet{yu2017learning}, we build a hybrid dialog system by combining the BlenderBot \cite{roller2020recipes} and our system. Since there is no intent label in this work, we use the prediction of the system attribute predictor as a proxy for selecting a chitchat content or an on-task content. When the output of the system attribute predictor is empty, we choose the response from the BlenderBot baseline. Otherwise, we choose the response from our response generator. We compare \textsc{Deux} with this model to explore the ability of generating social content in our response generator.

\paragraph{\textsc{Deux}-Usr} This is our ablated model where we only consider the attributes from user utterances in both attribute tracking module and system attribute predictor module. We build this model to examine contributions of considering attributes in previous system responses.


\subsection{Implementation Details}
We implement our attribute tracking module on pre-trained language model BERT \cite{devlin2019bert} and add the placeholders in \textsc{Inspired} as special tokens. We truncate input dialog history to 512 tokens. Our system attribute predictor module is developed upon medium size GPT-2 \cite{radford2019language}. Except for the same placeholders tokens, we also add a set of special tokens indicating new attributes in the next system attribute tracking. Sequences longer than 1024 tokens are truncated in system attribute predictor. Experiments for our approach use default hyperparameters for BERT and GPT-2 in Huggingface Transformers \cite{wolf2020transformers}. Our response generator and all the baselines use 2.7B BlenderBot model \cite{roller2020recipes}, we fine-tune BlenderBot with Adafactor \cite{shazeer2018adafactor} and conduct nucleus sampling during testing. All the models are implemented on two NVIDIA titan RTX GPUs. Since we mainly focus on the dialog systems, we use the same recommender system with \textsc{Inspired} Bot baseline in all our dialog systems. We use ParlAI framework \cite{miller2017parlai} to implement our code for building the model. 

\begin{table*}[htb!]
    \centering
    \begin{tabular}{l|ccccc}
    \hline
    \textbf{Model}  & \textbf{Consistency} & \textbf{Naturalness} & \textbf{ Engagingness} & \textbf{Sociability} & \textbf{Length} \\
    \hline
    \textsc{Inspired} Bot     & 3.40 &  3.29** & 3.51 & 3.74 & 8.64	\\
    BlenderBot   & 3.40 & 3.55 & 3.67 &  \textbf{3.86} & 	10.86 \\
    Hybrid     &  3.43 & 3.55 &  3.53 &  3.72 & 	9.62\\
    \textsc{Deux}-Usr     & 3.25* & 3.43**  & 3.34** &  3.69 & 9.69 \\ 
    \hline
    \textbf{\textsc{Deux}}     &  \textbf{3.64} & \textbf{3.79} & \textbf{3.78} & \textbf{3.86} & \textbf{11.91}	\\
    \hline
    \end{tabular}
    \caption{Results from human evaluation; we test all the baseline bots against \textsc{Deux} with **$p < 0.01$, *$p < 0.05$.}
    \label{tab:human_eval}
\end{table*}

\subsection{Metrics}
We adopt automatic evaluation for separate modules and human evaluation for the whole system. We have more number of human evaluation metrics than automatic metrics since sociable chat is a subjective task. We choose to use perplexity and token accuracy which are the default automatic metrics in ParlAI.

\paragraph{Automatic Metrics}
We calculate the accuracy and F1 scores of the placeholders in the predicted next system attribute tracking to measure the performance of attribute tracking predictors, we also calculate the per-token accuracy of the placeholders in the system attribute tracking\footnote{These metrics are the accuracy of placeholders instead of attributes because we can fill in the placeholders with attributes by any recommender system.}.
To evaluate the performance of the response generator, we report perplexity which measures the system response quality.

\paragraph{Human Evaluation} 
In most previous works of recommendation dialog systems, human evaluation often consists of assessing the utterances generated by the systems, e.g. in terms of their consistency with previous dialog history utterances. Such assessment may not be sufficient to show the practical usefulness of the recommendation dialog system. \citet{Jannach2020EndtoEndLF} found that about one third of the generated responses are not meaningful in the given context. To alleviate this problem in the human evaluation, we evaluate our system and other baselines with crowd-workers on Amazon Mechanical Turk.

We hire 58 workers on Amazon Mechanical Turk platform; each worker is assigned the task to be a movie seeker and interact with all the chatbots to get movie recommendation. Workers are required to use consistent movie preference to interact with all the chatbots. We randomize the order of the chatbots to avoid unfair evaluation. After chatting with each bot, workers are required to finish a post-survey and score each chatbot on four metrics in a 5-point Likert scale: 

\textbf{Consistency}: focuses more on the logical consistency between the system response and the dialog history.

\textbf{Naturalness}: Unlike consistency, naturalness is used to explore how human-like the systems' language generation quality. 

\textbf{Engagingness}: One goal of the sociable movie recommendation system is to keep engaging with users. Here, we evaluate if the user would like to continue chatting with the system.

\textbf{Sociability}: Our goal in this work is to combine on-task content and social content in the movie recommendation task. Sociability is used to evaluate if the bot has good social content in the responses and if the user considers the bot as friendly.

    
    
    

We also report the dialog length and the success rate of good recommendation. In the post-task survey, we ask the users if the chatbot recommends a movie and if the recommended movie fits their preference. To guarantee the quality of our human evaluation result, we ask these same questions before and after the task to filter out bad workers. If the answers are different, we drop the results. We provide more details of human evaluation in \autoref{appendix:eval_interface}.

\subsection{Results and Discussion}

\begin{table}[htb!]
\centering
\begin{tabular}{l|ccc}
    \hline
    \textbf{Model}& \textbf{Token Acc.} & \textbf{Acc.} & \textbf{F1}\\
    \hline
    \hline
    \textsc{Deux}-Usr & 0.55 & 0.34  & 0.46\\
    \hline
    \textbf{\textsc{Deux}} & \textbf{0.78} & \textbf{0.48}  & \textbf{0.63}\\
    \hline
\end{tabular}
\caption{Automatic evaluation results of attribute tracking predictor. \textbf{Token Acc.}: token accuracy of system attribute tracking \& \textbf{Acc.}: accuracy of system attribute tracking}
\label{tab:auto_eval_entity_tracking_predictor}
\end{table}

\begin{table}[htb!]
\centering
\begin{tabular}{l|cc}
    \hline
    \textbf{Model}& \textbf{Perplexity$\downarrow$} \\
    \hline
    \hline
    \textsc{Inspired} Bot & 8.93 \\
    BlenderBot & 8.95 \\ \hline
    \textbf{\textsc{Deux}} & \textbf{7.42}  \\
    \hline
\end{tabular}
\caption{Automatic evaluation results of response generator. The automatic evaluation results of \textsc{Inspired} Bot is from \citet{hayati2020inspired}.
  }
\label{tab:auto_eval_response_generator}
\vspace{-5mm}
\end{table}

The result from automatic metrics are shown in  Table \ref{tab:auto_eval_entity_tracking_predictor} and Table \ref{tab:auto_eval_response_generator}. Table \ref{tab:auto_eval_entity_tracking_predictor} presents the results of two different attribute tracking predictors. We observe that our attribute tracking predictor outperforms the baseline which only considers the attributes in user utterances on all the metrics. This result shows that in the movie recommendation task, tracking the attributes in previous system responses can contribute to the improvement of the system attribute tracking module's performance. From Table \ref{tab:auto_eval_response_generator}, we observe that \textsc{Deux} achieves a lower perplexity, indicating that our method has a better generation quality compared to \textsc{Inspired} Bot and BlenderBot. This indicates that using the attribute information in the response generator improves the quality of the system response.


For our human evaluations studies, 58 users participate in our user study. Each user chats with all chatbots. Table \ref{tab:human_eval} shows human evaluation results of different recommendation chatbots. We can see that \textsc{Deux} outperforms all the baselines on all metrics. Users rate \textsc{Deux} as the most consistent and natural bot, indicating it has better response quality than other baselines. \textsc{Deux} also maintains longer and more engaging conversations. This suggests that users are more willing to chat with \textsc{Deux} for a longer time. \textsc{Deux} also has a sociability score as high as the BlenderBot; this indicates that \textsc{Deux} contains diverse social interactions.

\begin{table*}[htb!]
    \centering
    \small
    \begin{tabular}{ll}
    \hline
    \hspace{2mm}(A) \textcolor{red}{Usr:} User \ \ \ \ \ \ \   \textcolor{blue}{Sys:} BlenderBot & \hspace{2mm}(B) \textcolor{red}{Usr:} User \ \ \ \ \ \ \   \textcolor{blue}{Sys:} \textsc{Deux}\hspace{70mm}\\
    \hline
    \begin{tabular}{lp{60mm}}
        \textcolor{blue}{Sys}: & Hello! How are you doing?\\
        \textcolor{red}{Usr:} & Do you know any \textbf{animated movies}? \\
        \textcolor{blue}{Sys:} & Yes, I am a fan of \textbf{cartoons}. \\
        \textcolor{red}{Usr:} & I enjoy \textbf{animated movies} from Japan.\\
        &...\\
        \textcolor{blue}{Sys:} & \textbf{\textbf{Alladin}} came out last year and it was a good movie.\\
        \textcolor{red}{Usr:} & Do you know any other movies?\\
        \textcolor{blue}{Sys:} & Yes i do. what kind of movies do you enjoy?\\
    \end{tabular}
    &
    \begin{tabular}{lp{60mm}}
        &...\\
        \textcolor{blue}{Sys:} & What kind of movies do you enjoy watching? \\
        \textcolor{red}{Usr:} & I really like \textbf{\textbf{horror}}, but I was thinking for tonight with the kids maybe an \textbf{\textbf{action movie}}?\\
        &...\\
        \textcolor{blue}{Sys:} & I would like to recommend \textbf{\textbf{Avengers: Endgame}}.\\
        \textcolor{red}{Usr:} & You know what, we have seen it. \\
        \textcolor{blue}{Sys:} & That is a great choice. have you watched any of the other \textbf{action movies} yet?\\
    \end{tabular}\\

    \hline
    \hspace{2mm}(C) \textcolor{red}{Usr:} User \ \ \ \ \ \ \   \textcolor{blue}{Sys}: \textsc{Deux}-Usr & \hspace{2mm}(D) \textcolor{red}{Usr:} User \ \ \ \ \ \ \   \textcolor{blue}{Sys}: \textsc{Deux}\hspace{70mm}\\
    \hline

    \begin{tabular}{lp{60mm}}
        &...\\
        \textcolor{red}{Usr:} & Who is your favorite actress?\\
        \textcolor{blue}{Sys:} & I really like \textbf{Sandra Bullock}\\
        \textcolor{red}{Usr:} & Can you recommend a movie with her? \\
        \textcolor{blue}{Sys:} & I would like to recommend \textbf{Like a Boss}? \\ 
        \textcolor{red}{Usr:} & She is not in that. \\
        \textcolor{blue}{Sys:} & What type of movies do you like?\\
        \textcolor{red}{Usr:} & Movies staring \textbf{Sandra Bullock}. \\
        \textcolor{blue}{Sys:} & I am recommending: \textbf{Gravity}\\
    \end{tabular}
    &
    \begin{tabular}{lp{60mm}}
        &...\\
        \textcolor{red}{Usr:} & I like \textbf{horror movie} \\
        \textcolor{blue}{Sys:} & I recommend the movie \textbf{Antlers} \\
        \textcolor{red}{Usr:} & I have not heard of that, who is the star?\\
        \textcolor{blue}{Sys:} & \textbf{Keri Russell}\\
        \textcolor{red}{Usr:} & Is she a new actress? I haven't heard of her. \\
        \textcolor{blue}{Sys:} & Have you heard about \textbf{Tom Cruise}? \\ 
        \textcolor{red}{Usr:} & Yes of course. \\
        \textcolor{blue}{Sys:} & They were cast together for \textbf{Mission Impossible}.\\
    \end{tabular}\\
    \hline
    \end{tabular}
\caption{
Examples of human-machine dialogs in human evaluation. Attributes in the utterances are bolded.
}
\label{table:case study}
\vspace{-2mm}
\end{table*}

Compared to \textsc{Deux}-Usr which only tracks attributes in user utterances, \textsc{Deux} receives significantly higher average consistency score. It suggests that system responses in \textsc{Deux} are more consistent with the dialog history since it also tracks the attributes in previous system responses. We also notice that \textsc{Deux} has significantly higher engagingness score than \textsc{Deux}-Usr. It indicates that the attribute tracking module in the system side improves the sociability of the bot. 


\begin{figure}[t!]
\vspace{-2mm}
\centering
{
\includegraphics[trim={0 0.8cm 0 0.8cm},clip,width=7.7cm]{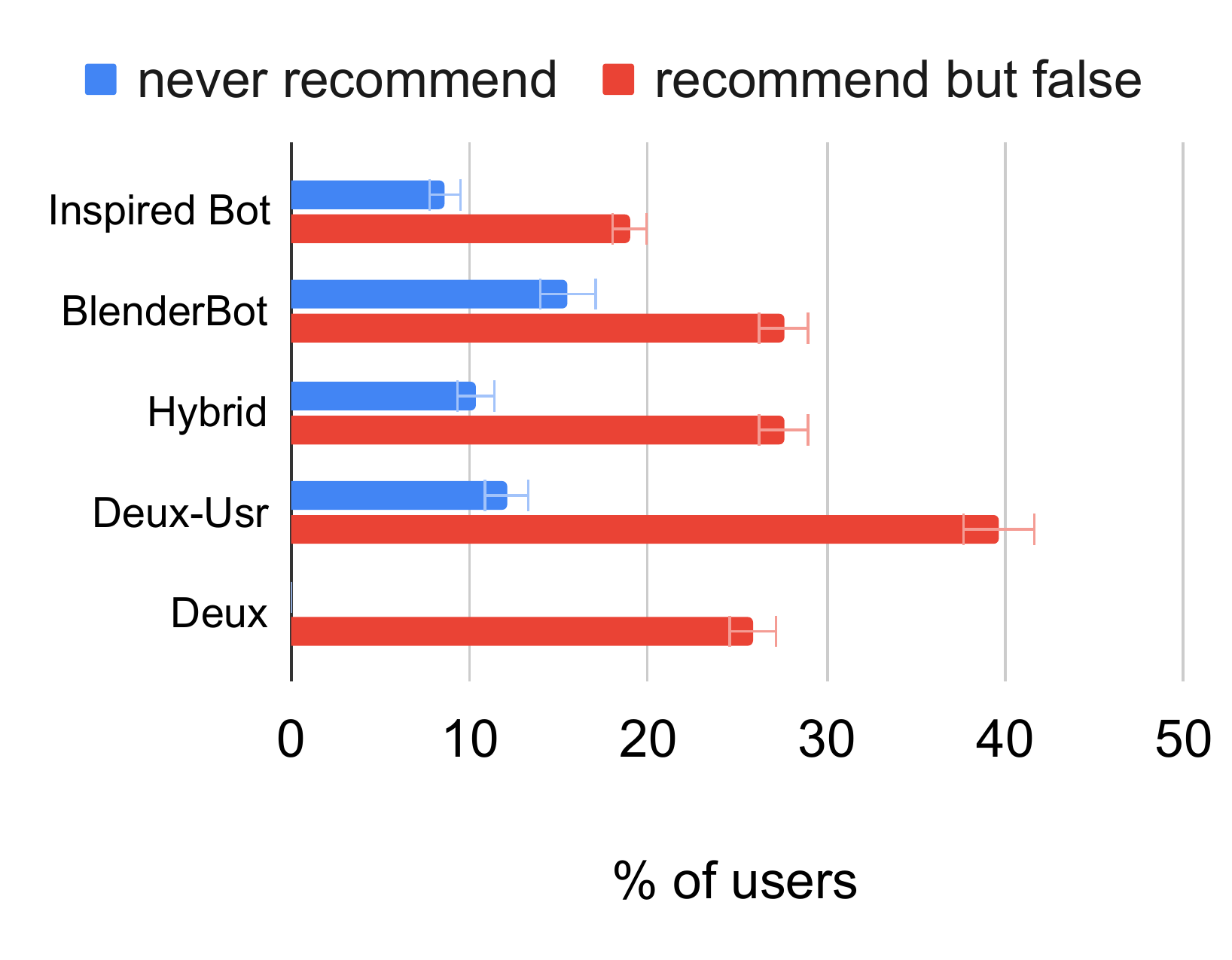}
}
\caption{\label{fig:task_result} Percentage of users who reported in the post-task survey that the bot never recommends a movie or the bot recommends a movie but it does not fit the user preferences.}
\vspace{-5mm}
\end{figure}


If we look more into the details of the task success in Figure \ref{fig:task_result}, \textsc{Deux} always recommends a movie compared to other baseline bots. BlenderBot has the highest number of users (27.59\%) who report that the bot never recommends a movie. Then, if we only consider the user-side attribute tracking \
\textsc{Deux}-Usr, it does recommend movies to the users, but mostly users report that the movies do not suit their taste (39.66\%, compared to \textsc{Deux} which only has 25.86\%). Thus, it shows that the system-side attribute tracking in \textsc{Deux} considers the attributes introduced by the system and helps the bot to find movies that fit the user preferences best. 

\subsection{Analyses of Human-Machine Dialogs}
We show four examples of human-machine dialogs from our human evaluation in Table \ref{table:case study}. First, we compare the conversations generated by the BlenderBot baseline (Ex. \ref{table:case study}A) and the conversations generated by  \textsc{Deux} (Ex. \ref{table:case study}B) to explore the influence of our attribute tracking module. In Example \ref{table:case study}A, we can see how the user mentions that ``animated movies'' as their preferred movie genre, rejects the first recommendation, and requests for another one. Then, the BlenderBot is requesting the user's movie preference again (the last line in Ex. \ref{table:case study}A) even though it should have provided another animated movie. This is because the BlenderBot doesn't track the attributes in the dialog so that it can ask repeated questions. 

In Example \ref{table:case study}B, \textsc{Deux} has similar context where the user asks for an ``action movie'' and refuses the first recommendation. \textsc{Deux} extracts action movies instead of horror movies as the positive attribute and predicts the ``action movies'' to be in the next system attribute to track. Then, \textsc{Deux} asks the user about what action movies the user has watched, which is consistent with the history. 

Now, let us look at the conversations with \textsc{Deux}-Usr (Ex. \ref{table:case study}C) and another conversation with \textsc{Deux} (Ex. \ref{table:case study}D) to explore if tracking the attributes at the system side helps. In Example \ref{table:case study}C, the system says that ``Sandra Bullock'' is its favorite actress. However, it recommends a movie without this actress in the following turn since it does not track the attributes in its own response. After the user mentions the same actress again in the user utterance, the system recommends an appropriate movie (``Gravity'') with ``Sandra Bullock'' in it. Meanwhile, in Example \ref{table:case study}D, there are three attributes in previous responses. \textsc{Deux} tracks that ``Keri Russell'' is acting in ``Antlers''. Later, it suggests an actor, ``Tom Cruise'', and then generates a natural response which recommends a movie (``Mission Impossible'') where both ``Tom Cruise'' and ``Keri Russell'' starred. These observations indicate the contribution of system attribute tracking in recommendation dialog systems for a consistent response.

\section{Conclusion and Future Work}
In this paper, we present \textsc{Deux}, a novel attribute-guided framework for sociable recommendation dialog systems. It tracks attributes in the conversation from both user's and system's sides. Thus, our framework can generate more consistent and natural responses compared to previous methods. While \textsc{Deux} can be applied to any recommendation dialog tasks, we evaluate it on a movie recommendation dataset. \textsc{Deux} outperforms the existing dialog systems in both automatic metrics and human evaluation.

Looking forward, a future direction would be to incorporate the dialog policies or strategies in our attribute-guided framework. By integrating the strategies with our attribute tracking slots, we hope that the dialog system can produce better social content and task-oriented content. Another interesting direction would be to expand the framework to consider other attributes of the recommended item, such as the movie plots which have more complex characteristics compared to current attributes.


\bibliography{anthology,custom}

\begin{thebibliography}{32}
\expandafter\ifx\csname natexlab\endcsname\relax\def\natexlab#1{#1}\fi

\bibitem[{Budzianowski et~al.(2020)Budzianowski, Wen, Tseng, Casanueva, Ultes,
  Ramadan, and Gašić}]{budzianowski2020multiwoz}
Paweł Budzianowski, Tsung-Hsien Wen, Bo-Hsiang Tseng, Iñigo Casanueva, Stefan
  Ultes, Osman Ramadan, and Milica Gašić. 2020.
\newblock \href {http://arxiv.org/abs/1810.00278} {Multiwoz -- a large-scale
  multi-domain wizard-of-oz dataset for task-oriented dialogue modelling}.

\bibitem[{Chen et~al.(2019)Chen, Lin, Zhang, Ding, Cen, Yang, and
  Tang}]{chen-etal-2019-towards}
Qibin Chen, Junyang Lin, Yichang Zhang, Ming Ding, Yukuo Cen, Hongxia Yang, and
  Jie Tang. 2019.
\newblock \href {https://doi.org/10.18653/v1/D19-1189} {Towards knowledge-based
  recommender dialog system}.
\newblock In \emph{Proceedings of the 2019 Conference on Empirical Methods in
  Natural Language Processing and the 9th International Joint Conference on
  Natural Language Processing (EMNLP-IJCNLP)}, pages 1803--1813, Hong Kong,
  China. Association for Computational Linguistics.

\bibitem[{Christakopoulou et~al.(2018)Christakopoulou, Beutel, Li, Jain, and
  hsin Chi}]{christakopoulou2018q}
Konstantina Christakopoulou, Alex Beutel, R.~Li, S.~Jain, and Ed~Huai hsin Chi.
  2018.
\newblock Q\&r: A two-stage approach toward interactive recommendation.
\newblock \emph{Proceedings of the 24th ACM SIGKDD International Conference on
  Knowledge Discovery \& Data Mining}.

\bibitem[{Devlin et~al.(2019)Devlin, Chang, Lee, and
  Toutanova}]{devlin2019bert}
Jacob Devlin, Ming-Wei Chang, Kenton Lee, and Kristina Toutanova. 2019.
\newblock \href {http://arxiv.org/abs/1810.04805} {Bert: Pre-training of deep
  bidirectional transformers for language understanding}.

\bibitem[{Hayati et~al.(2020)Hayati, Kang, Zhu, Shi, and
  Yu}]{hayati2020inspired}
Shirley~Anugrah Hayati, Dongyeop Kang, Qingxiaoyang Zhu, Weiyan Shi, and Zhou
  Yu. 2020.
\newblock \href {https://www.aclweb.org/anthology/2020.emnlp-main.654}
  {{INSPIRED}: Toward sociable recommendation dialog systems}.
\newblock In \emph{Proceedings of the 2020 Conference on Empirical Methods in
  Natural Language Processing (EMNLP)}, pages 8142--8152, Online. Association
  for Computational Linguistics.

\bibitem[{Henderson et~al.(2013)Henderson, Thomson, and
  Young}]{henderson2013deep}
Matthew Henderson, B.~Thomson, and S.~Young. 2013.
\newblock Deep neural network approach for the dialog state tracking challenge.
\newblock In \emph{SIGDIAL Conference}.

\bibitem[{Hosseini-Asl et~al.(2020)Hosseini-Asl, McCann, Wu, Yavuz, and
  Socher}]{hosseini2020simple}
Ehsan Hosseini-Asl, Bryan McCann, Chien-Sheng Wu, Semih Yavuz, and Richard
  Socher. 2020.
\newblock \href {http://arxiv.org/abs/2005.00796} {A simple language model for
  task-oriented dialogue}.

\bibitem[{Jannach and Manzoor(2020)}]{Jannach2020EndtoEndLF}
D.~Jannach and Ahtsham Manzoor. 2020.
\newblock End-to-end learning for conversational recommendation: A long way to
  go?
\newblock In \emph{IntRS@RecSys}.

\bibitem[{Kang et~al.(2019)Kang, Balakrishnan, Shah, Crook, Boureau, and
  Weston}]{kang2019-recommendation}
Dongyeop Kang, Anusha Balakrishnan, Pararth Shah, Paul Crook, Y-Lan Boureau,
  and Jason Weston. 2019.
\newblock \href {https://doi.org/10.18653/v1/D19-1203} {Recommendation as a
  communication game: Self-supervised bot-play for goal-oriented dialogue}.
\newblock In \emph{Proceedings of the 2019 Conference on Empirical Methods in
  Natural Language Processing and the 9th International Joint Conference on
  Natural Language Processing (EMNLP-IJCNLP)}, pages 1951--1961, Hong Kong,
  China. Association for Computational Linguistics.

\bibitem[{Lee et~al.(2018)Lee, Moore, Ren, Arar, and Jiang}]{lee2018making}
Sunhwan Lee, Robert Moore, Guang-Jie Ren, Raphael Arar, and Shun Jiang. 2018.
\newblock Making personalized recommendation through conversation: Architecture
  design and recommendation methods.
\newblock In \emph{Workshops at the Thirty-Second AAAI Conference on Artificial
  Intelligence}.

\bibitem[{Li et~al.(2019{\natexlab{a}})Li, Kahou, Schulz, Michalski, Charlin,
  and Pal}]{li2018towards}
Raymond Li, Samira Kahou, Hannes Schulz, Vincent Michalski, Laurent Charlin,
  and Chris Pal. 2019{\natexlab{a}}.
\newblock \href {http://arxiv.org/abs/1812.07617} {Towards deep conversational
  recommendations}.

\bibitem[{Li et~al.(2019{\natexlab{b}})Li, Qian, Shi, and Yu}]{li2020end}
Yu~Li, Kun Qian, Weiyan Shi, and Zhou Yu. 2019{\natexlab{b}}.
\newblock \href {http://arxiv.org/abs/1911.10742} {End-to-end trainable
  non-collaborative dialog system}.

\bibitem[{Liang et~al.(2020)Liang, Chau, Li, Lu, Yu, Zhou, Jain, Davidson,
  Arnold, Nguyen, and Yu}]{gunrock2020}
Kaihui Liang, Austin Chau, Yu~Li, Xueyuan Lu, Dian Yu, Mingyang Zhou, Ishan
  Jain, Sam Davidson, Josh Arnold, Minh Nguyen, and Zhou Yu. 2020.
\newblock \href
  {https://m.media-amazon.com/images/G/01/mobile-apps/dex/alexa/alexaprize/assets/challenge3/proceedings/UC-Davis-Gunrock.pdf}
  {Gunrock 2.0: A user adaptive social conversational system}.
\newblock In \emph{3rd Proceedings of Alexa Prize (Alexa Prize 2020)}.

\bibitem[{Liu et~al.(2019)Liu, Ott, Goyal, Du, Joshi, Chen, Levy, Lewis,
  Zettlemoyer, and Stoyanov}]{liu2019roberta}
Yinhan Liu, Myle Ott, Naman Goyal, Jingfei Du, Mandar Joshi, Danqi Chen, Omer
  Levy, Mike Lewis, Luke Zettlemoyer, and Veselin Stoyanov. 2019.
\newblock \href {http://arxiv.org/abs/1907.11692} {Roberta: A robustly
  optimized bert pretraining approach}.

\bibitem[{Liu et~al.(2020)Liu, Wang, Niu, Wu, Che, and
  Liu}]{liu-etal-2020-towards}
Zeming Liu, Haifeng Wang, Zheng-Yu Niu, Hua Wu, Wanxiang Che, and Ting Liu.
  2020.
\newblock \href {https://doi.org/10.18653/v1/2020.acl-main.98} {Towards
  conversational recommendation over multi-type dialogs}.
\newblock In \emph{Proceedings of the 58th Annual Meeting of the Association
  for Computational Linguistics}, pages 1036--1049, Online. Association for
  Computational Linguistics.

\bibitem[{Miller et~al.(2018)Miller, Feng, Fisch, Lu, Batra, Bordes, Parikh,
  and Weston}]{miller2017parlai}
Alexander~H. Miller, Will Feng, Adam Fisch, Jiasen Lu, Dhruv Batra, Antoine
  Bordes, Devi Parikh, and Jason Weston. 2018.
\newblock \href {http://arxiv.org/abs/1705.06476} {Parlai: A dialog research
  software platform}.

\bibitem[{Moon et~al.(2019)Moon, Shah, Kumar, and Subba}]{moon2019opendialkg}
Seungwhan Moon, Pararth Shah, Anuj Kumar, and Rajen Subba. 2019.
\newblock \href {https://doi.org/10.18653/v1/P19-1081} {{O}pen{D}ial{KG}:
  Explainable conversational reasoning with attention-based walks over
  knowledge graphs}.
\newblock In \emph{Proceedings of the 57th Annual Meeting of the Association
  for Computational Linguistics}, pages 845--854, Florence, Italy. Association
  for Computational Linguistics.

\bibitem[{Mrk{\v{s}}i{\'c} et~al.(2017)Mrk{\v{s}}i{\'c}, {\'O}~S{\'e}aghdha,
  Wen, Thomson, and Young}]{mrkvsic2016neural}
Nikola Mrk{\v{s}}i{\'c}, Diarmuid {\'O}~S{\'e}aghdha, Tsung-Hsien Wen, Blaise
  Thomson, and Steve Young. 2017.
\newblock \href {https://doi.org/10.18653/v1/P17-1163} {Neural belief tracker:
  Data-driven dialogue state tracking}.
\newblock In \emph{Proceedings of the 55th Annual Meeting of the Association
  for Computational Linguistics (Volume 1: Long Papers)}, pages 1777--1788,
  Vancouver, Canada. Association for Computational Linguistics.

\bibitem[{Papaioannou et~al.(2017)Papaioannou, Dondrup, Novikova, and
  Lemon}]{papaioannou2017hybrid}
Ioannis Papaioannou, Christian Dondrup, Jekaterina Novikova, and Oliver Lemon.
  2017.
\newblock \href {https://doi.org/10.1109/ROMAN.2017.8172363} {Hybrid chat and
  task dialogue for more engaging hri using reinforcement learning}.
\newblock In \emph{2017 26th IEEE International Symposium on Robot and Human
  Interactive Communication (RO-MAN)}, IEEE International Symposium on Robot
  and Human Interactive Communication, pages 593--598, United States. IEEE.
\newblock 26th IEEE International Symposium on Robot and Human Interactive
  Communication 2017, RO-MAN 2017 ; Conference date: 28-08-2017 Through
  01-09-2017.

\bibitem[{Radford et~al.(2019)Radford, Wu, Child, Luan, Amodei, and
  Sutskever}]{radford2019language}
Alec Radford, Jeff Wu, Rewon Child, David Luan, Dario Amodei, and Ilya
  Sutskever. 2019.
\newblock Language models are unsupervised multitask learners.

\bibitem[{Rastogi et~al.(2018)Rastogi, Hakkani-Tur, and
  Heck}]{rastogi2017scalable}
Abhinav Rastogi, Dilek Hakkani-Tur, and Larry Heck. 2018.
\newblock \href {http://arxiv.org/abs/1712.10224} {Scalable multi-domain
  dialogue state tracking}.

\bibitem[{Roller et~al.(2020)Roller, Dinan, Goyal, Ju, Williamson, Liu, Xu,
  Ott, Shuster, Smith, Boureau, and Weston}]{roller2020recipes}
Stephen Roller, Emily Dinan, Naman Goyal, Da~Ju, Mary Williamson, Yinhan Liu,
  Jing Xu, Myle Ott, Kurt Shuster, Eric~M. Smith, Y-Lan Boureau, and Jason
  Weston. 2020.
\newblock \href {http://arxiv.org/abs/2004.13637} {Recipes for building an
  open-domain chatbot}.

\bibitem[{Shazeer and Stern(2018)}]{shazeer2018adafactor}
Noam Shazeer and Mitchell Stern. 2018.
\newblock \href {http://proceedings.mlr.press/v80/shazeer18a.html} {Adafactor:
  Adaptive learning rates with sublinear memory cost}.
\newblock In \emph{Proceedings of the 35th International Conference on Machine
  Learning}, volume~80 of \emph{Proceedings of Machine Learning Research},
  pages 4596--4604. PMLR.

\bibitem[{Sun and Zhang(2018)}]{sun2018conversational}
Yueming Sun and Yi~Zhang. 2018.
\newblock Conversational recommender system.
\newblock In \emph{The 41st International ACM SIGIR Conference on Research \&
  Development in Information Retrieval}, pages 235--244.

\bibitem[{Vaswani et~al.(2017)Vaswani, Shazeer, Parmar, Uszkoreit, Jones,
  Gomez, Kaiser, and Polosukhin}]{vaswani2017attention}
Ashish Vaswani, Noam Shazeer, Niki Parmar, Jakob Uszkoreit, Llion Jones,
  Aidan~N. Gomez, Lukasz Kaiser, and Illia Polosukhin. 2017.
\newblock \href {http://arxiv.org/abs/1706.03762} {Attention is all you need}.

\bibitem[{Wolf et~al.(2020)Wolf, Debut, Sanh, Chaumond, Delangue, Moi, Cistac,
  Rault, Louf, Funtowicz, Davison, Shleifer, von Platen, Ma, Jernite, Plu, Xu,
  Le~Scao, Gugger, Drame, Lhoest, and Rush}]{wolf2020transformers}
Thomas Wolf, Lysandre Debut, Victor Sanh, Julien Chaumond, Clement Delangue,
  Anthony Moi, Pierric Cistac, Tim Rault, Remi Louf, Morgan Funtowicz, Joe
  Davison, Sam Shleifer, Patrick von Platen, Clara Ma, Yacine Jernite, Julien
  Plu, Canwen Xu, Teven Le~Scao, Sylvain Gugger, Mariama Drame, Quentin Lhoest,
  and Alexander Rush. 2020.
\newblock \href {https://doi.org/10.18653/v1/2020.emnlp-demos.6} {Transformers:
  State-of-the-art natural language processing}.
\newblock In \emph{Proceedings of the 2020 Conference on Empirical Methods in
  Natural Language Processing: System Demonstrations}, pages 38--45, Online.
  Association for Computational Linguistics.

\bibitem[{Wu et~al.(2019)Wu, Zhang, Li, and Yu}]{wu2019alternating}
Qingyang Wu, Yichi Zhang, Yu~Li, and Zhou Yu. 2019.
\newblock \href {http://arxiv.org/abs/1910.03756} {Alternating roles dialog
  model with large-scale pre-trained language models}.

\bibitem[{Yang et~al.(2020)Yang, Dai, Yang, Carbonell, Salakhutdinov, and
  Le}]{yang2020xlnet}
Zhilin Yang, Zihang Dai, Yiming Yang, Jaime Carbonell, Ruslan Salakhutdinov,
  and Quoc~V. Le. 2020.
\newblock \href {http://arxiv.org/abs/1906.08237} {Xlnet: Generalized
  autoregressive pretraining for language understanding}.

\bibitem[{Yu et~al.(2017)Yu, Black, and Rudnicky}]{yu2017learning}
Zhou Yu, Alan~W Black, and Alexander~I. Rudnicky. 2017.
\newblock \href {http://arxiv.org/abs/1703.00099} {Learning conversational
  systems that interleave task and non-task content}.

\bibitem[{Zhang et~al.(2018)Zhang, Chen, Ai, Yang, and
  Croft}]{zhang2018towards}
Yongfeng Zhang, Xu~Chen, Qingyao Ai, Liu Yang, and W~Bruce Croft. 2018.
\newblock Towards conversational search and recommendation: System ask, user
  respond.
\newblock In \emph{Proceedings of the 27th ACM International Conference on
  Information and Knowledge Management}, pages 177--186.

\bibitem[{Zhao et~al.(2018)Zhao, Romero, and Rudnicky}]{zhao2018sogo}
R.~Zhao, Oscar~J. Romero, and A.~Rudnicky. 2018.
\newblock Sogo: A social intelligent negotiation dialogue system.
\newblock \emph{Proceedings of the 18th International Conference on Intelligent
  Virtual Agents}.

\bibitem[{Zhou et~al.(2020)Zhou, Zhao, Bian, Zhou, Wen, and
  Yu}]{zhou2020improving}
Kun Zhou, Wayne~Xin Zhao, Shuqing Bian, Yuanhang Zhou, Ji-Rong Wen, and
  Jingsong Yu. 2020.
\newblock \href {http://arxiv.org/abs/2007.04032} {Improving conversational
  recommender systems via knowledge graph based semantic fusion}.

\end{thebibliography}
\bibliographystyle{acl_natbib}

\clearpage
\appendix

\section{Human Evaluation Details}
\label{appendix:eval_interface}
\begin{figure*}[htb!]
    \centering
    \includegraphics[trim={0 0 0 0},clip,width=16cm]{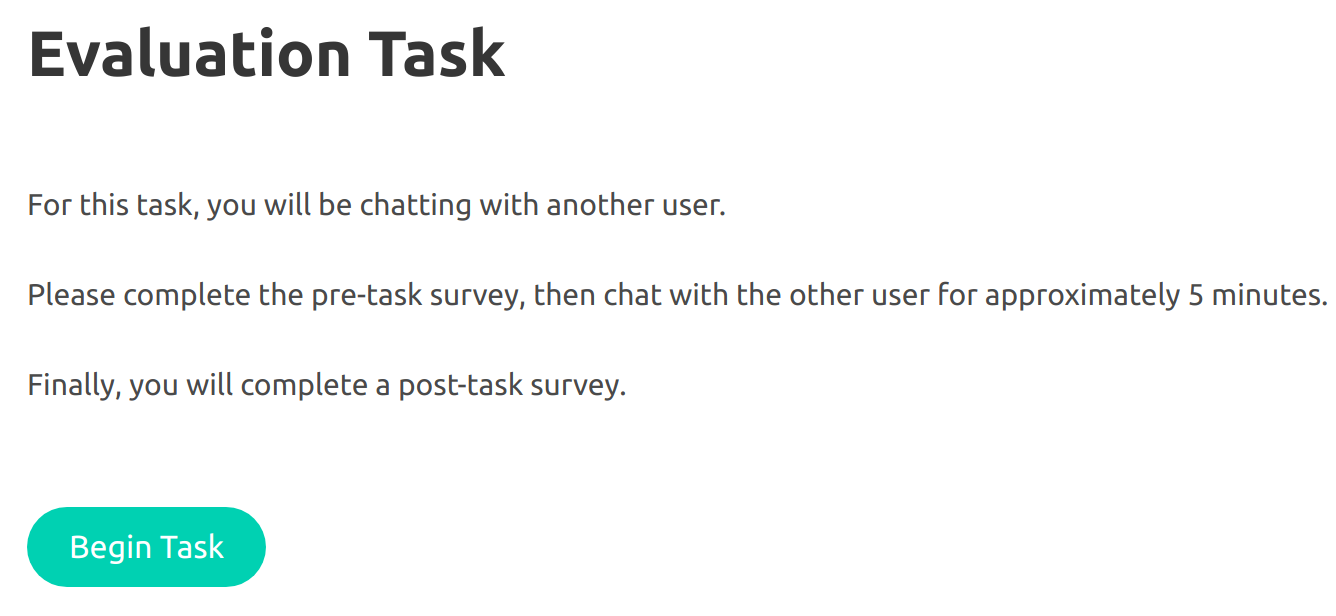}
    \caption{Screenshot of the human evaluation task instruction.}
    \label{fig:human_eval_1}
    \vspace{-3mm}
\end{figure*}

\begin{figure*}[htb!]
    \centering
    \includegraphics[trim={0 0 0 0},clip,width=16cm]{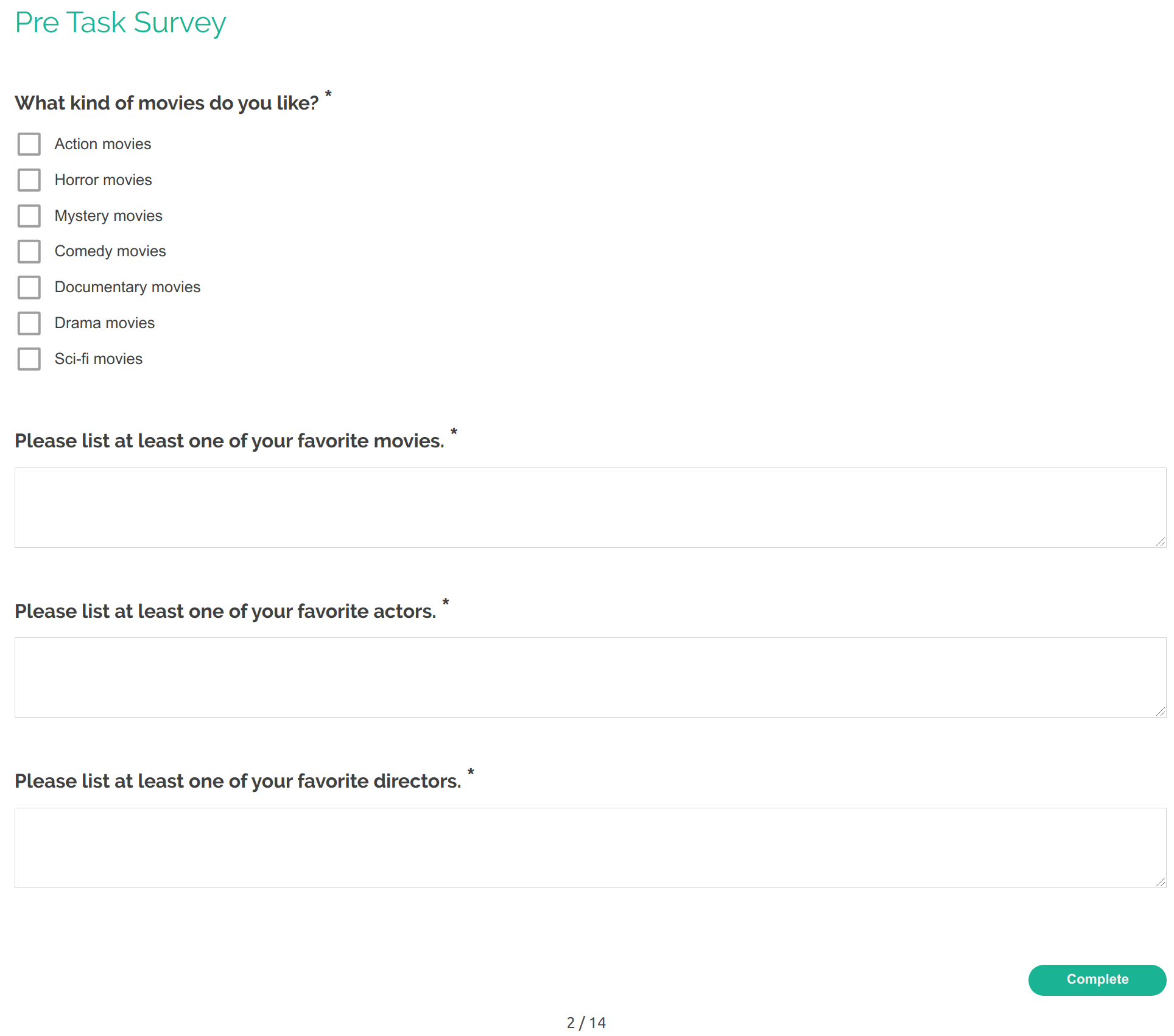}
    \caption{Screenshot of the pre-survey.}
    \label{fig:human_eval_2}
    \vspace{-3mm}
\end{figure*}

\begin{figure*}[htb!]
    \centering
    \includegraphics[trim={0 0 0 0},clip,width=16cm]{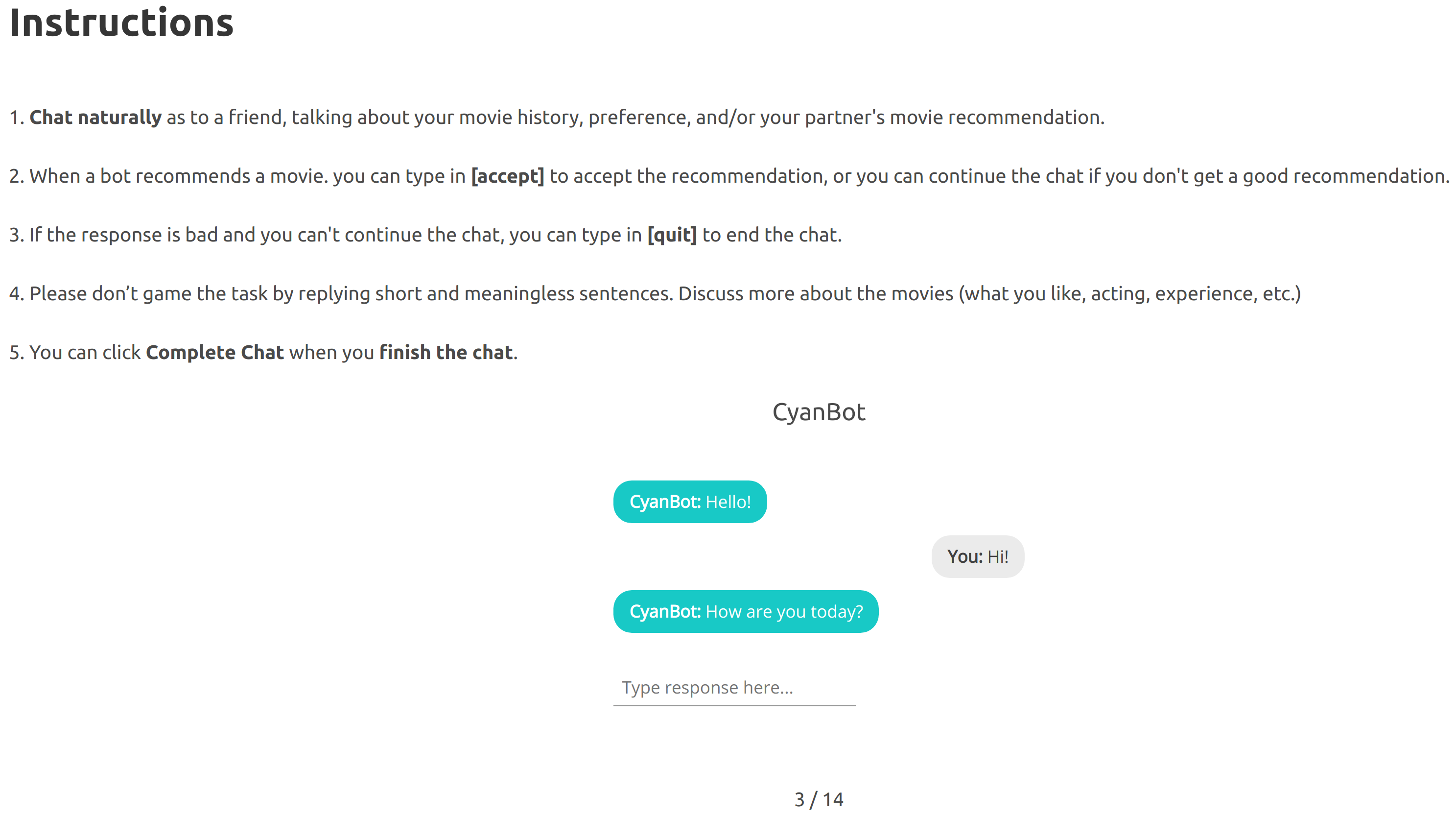}
    \caption{Screenshot of the interaction task interface.}
    \label{fig:human_eval_3}
    \vspace{-3mm}
\end{figure*}

\begin{figure*}[htb!]
    \centering
    \includegraphics[trim={0 0 0 0},clip,width=16cm]{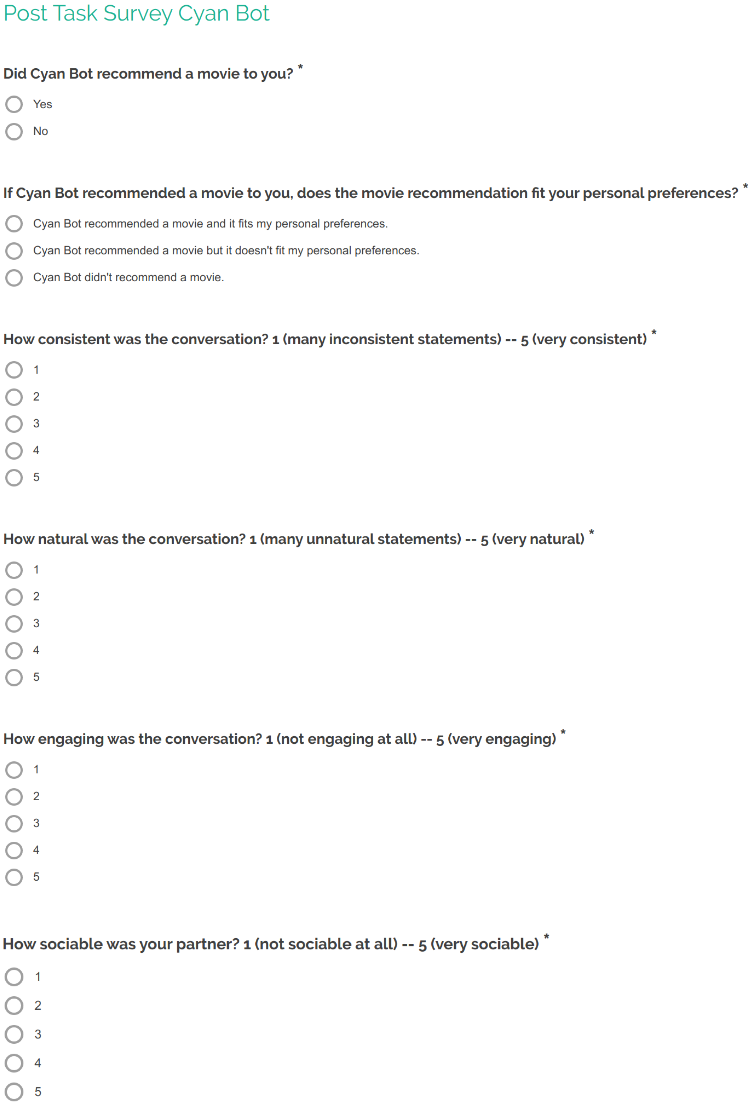}
    \caption{Screenshot of the survey after finishing every interaction task.}
    \label{fig:human_eval_4}
    \vspace{-3mm}
\end{figure*}

\begin{figure*}[htb!]
    \centering
    \includegraphics[trim={0 0 0 0},clip,width=16cm]{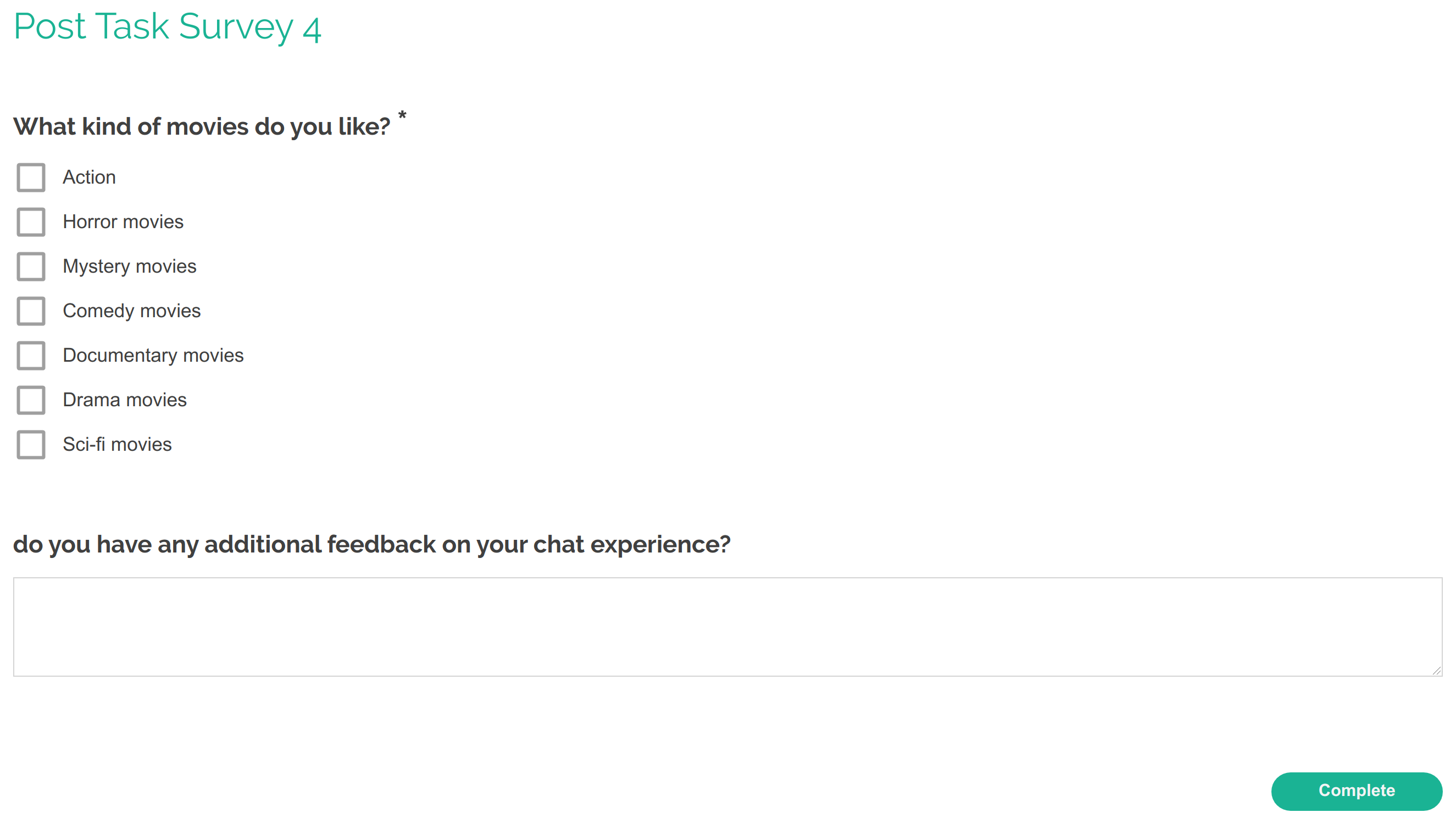}
    \caption{Screenshot of the post-survey.}
    \label{fig:human_eval_5}
    \vspace{-3mm}
\end{figure*}

Figure \ref{fig:human_eval_1} shows the instruction we give to the Amazon Mechanical Turkers in the human evaluation task. Figure \ref{fig:human_eval_2} shows the questions in the pre-survey, questions are designed for two purposes: first, these questions help Turkers think about related topics they may discuss in the movie recommendation conversations thus they can provide consistent answers in all the conversations. Second, the multiple choice question about movie genres is used to check the quality of the evaluation: we ask a same multiple choice question in the post-survey, if the answers are different, we think the Turker fails to provide consistent answers and remove the result. Figure \ref{fig:human_eval_3} shows the interface of the interaction task, we assign our system and other four baselines in random order, Turkers are asked to interact with one chatbot and finish the survey shown in Figure \ref{fig:human_eval_4} about the chatbot performance. Figure \ref{fig:human_eval_5} shows the post-survey questions, it includes a quality checking question mentioned above and an open question to collect additional feedback. Every Turker can only do the task once, we collect human evaluation results from 90 people on Mechanical Turk. We remove 32 Turkers who fail the quality check questions.

\end{document}